\documentclass[letterpaper, 10 pt, conference]{ieeeconf}  % Comment this line out if you need a4paper

\IEEEoverridecommandlockouts                              % This command is only needed if 
\usepackage{bm, amsmath,amssymb,mathtools}    % for formatting equations
\usepackage{pgfplots}
\usepackage{lipsum}
\usepackage{subcaption}
\usepackage{tabularx}
\usepackage{float}

\DeclareMathOperator*{\E}{\mathbb{E}} % short for expected value
\newtheorem{manualtheoreminner}{Boundary}
\newenvironment{manualtheorem}[1]{%
  \renewcommand\themanualtheoreminner{#1}%
  \manualtheoreminner
}{\endmanualtheoreminner}
\newcommand\inputpgf[2]{{
\let\pgfimageWithoutPath\pgfimage
\renewcommand{\pgfimage}[2][]{\pgfimageWithoutPath[##1]{#1/##2}}
\input{#1/#2}
}}

\DeclareMathOperator{\Blangle}{\Big\langle} % short for bigger inner product angle
\DeclareMathOperator{\Brangle}{\Big\rangle} % short for bigger inner product angle
\DeclareMathOperator{\Babs}{\Big|} % short for bigger abs
\DeclareMathOperator{\babs}{\big|} % short for bigger abs
\newtheorem{remark}{}
\newtheorem{lemma}{Lemma}

\newcommand\numberthis{\addtocounter{equation}{1}\tag{\theequation}} % allows numbering in align* env

\title{\LARGE \bf
Safe Continuous Control with Constrained Model-Based Policy Optimization}

\author{Moritz A. Zanger$^{*}$, Karam Daaboul$^{*}$ and J. Marius Z\"ollner% <-this % stops a space
\thanks{*Equal contributions}% <-this % stops a space
\thanks{Karlsruhe Institute of Technology, Kaiserstr. 12, 76131 Karlsruhe, Germany
        {\tt\small moritz.zanger@student.kit.edu, \{daaboul, marius.zoellner\}@kit.edu}}%
}

\begin{document}

\maketitle
\thispagestyle{empty}
\pagestyle{empty}

%%%%%%%%%%%%%%%%%%%%%%%%%%%%%%%%%%%%%%%%%%%%%%%%%%%%%%%%%%%%%%%%%%%%%%%%%%%%%%%%
\begin{abstract}
The applicability of reinforcement learning (RL) algorithms in real-world domains often requires adherence to safety constraints, a need difficult to address given the asymptotic nature of the classic RL optimization objective. In contrast to the traditional RL objective, safe exploration considers the maximization of expected returns under safety constraints expressed in expected cost returns. We introduce a model-based safe exploration algorithm for constrained high-dimensional control to address the often prohibitively high sample complexity of model-free safe exploration algorithms. Further, we provide theoretical and empirical analyses regarding the implications of model-usage on constrained policy optimization problems and introduce a practical algorithm that accelerates policy search with model-generated data. The need for accurate estimates of a policy's constraint satisfaction is in conflict with accumulating model-errors. We address this issue by quantifying model-uncertainty as the expected Kullback-Leibler divergence between predictions of an ensemble of probabilistic dynamics models and constrain this error-measure, resulting in an adaptive resampling scheme and dynamically limited rollout horizons. We evaluate this approach on several simulated constrained robot locomotion tasks with high-dimensional action- and state-spaces. Our empirical studies find that our algorithm reaches model-free performances with a 10-20 fold reduction of training samples while maintaining approximate constraint satisfaction levels of model-free methods. 
\end{abstract}

%%%%%%%%%%%%%%%%%%%%%%%%%%%%%%%%%%%%%%%%%%%%%%%%%%%%%%%%%%%%%%%%%%%%%%%%%%%%%%%%
\section{INTRODUCTION}
In the reinforcement learning (RL) paradigm, an agent aims to learn behavior from interactions with an initially unknown environment. The recent surge in computational capability and deep learning has aided substantial advancements in this field. Learning agents have demonstrated to be capable of producing complex behaviors ranging from continuous high-dimensional control gaits to cooperative tool usage\ \cite{Schulmanetal_ICLR2016,haarnojaSoftActorCriticOffPolicy2018a,fujimotoAddressingFunctionApproximation2018b,andrychowiczLearningDexterousInhand2020,bakerEmergentToolUse2020,martinez-gilEmergentBehaviorsScalability2017a}.

A characteristic inherent to canonical RL-algorithms is the need for a carefully designed reward function and numerous trials and errors to achieve well-performing policies. The designer's influence on an agent's behavior through reward specification is indirect and typically targets asymptotic performance rather than behavior during learning. In contrast, broad applicability in robot control often demands constraints on collisions, human interference, or other limits throughout the entire training process. Safe exploration methods aim at reconciling the conflict of exploring unknown (risky) state- and action-subspaces with safety constraints. In this work, we address constrained policy optimization problems, which consider safety constraints as a limit on a policy's expected cost returns rather than enforcing safety for individual trajectories\ \cite{rayBenchmarkingSafeExploration}. This problem poses the distinctive challenge of accurately estimating expected cost returns\ \cite{achiamConstrainedPolicyOptimization2017a}, often rendering algorithms prohibitively sample-inefficient and expensive to deploy to physical robots. 

To this end, model-based Reinforcement Learning (MBRL) has proven capable of significantly reducing sample complexity by leveraging a learned model of the system dynamics to augment policy learning with model-generated trajectories. However, model bias and accumulating prediction errors hinder accurate representations of the underlying state- and action-distributions, rendering their adaptation to expectation-based constrained policy search difficult. 

% Recursive state predictions, however, suffer from a potentially unbounded accumulation of errors. Previous approaches respond to this dilemma by limiting the task horizon\ \cite{kurutachModelEnsembleTrustRegionPolicy2018} or starting model-generated trajectories from buffered off-policy states, also referred to as branched rollouts\ \cite{jannerWhenTrustYour2019a}. Albeit effective for a number of tasks, both approaches change the underlying state-distribution, thus rendering their adaptation to expectation-based constrained policy search difficult. 

This work aims to improve the sample efficiency of safe exploration by reconciling model-usage with constrained policy search. We provide theoretical analysis on the implications of using imperfect synthetic data and derive a boundary on expected cost- and reward-changes due to model-based policy updates. Based on these results, we devise several measures to reconcile model-usage with guarantees on approximate safety constraint satisfaction in a practical algorithm. Specifically, these contributions involve 1.) a scale-invariant quantification of epistemic model-uncertainty with an ensemble of dynamics models, 2.) a dynamic resampling scheme to maintain an upper bound on average model errors and to prevent unsafe policy updates, 3.) an adaptive rollout scheme to contain accumulating model-errors through recursive successor-state predictions. 

We evaluate our algorithm on simulated high-dimensional continuous control tasks for robot locomotion with non-trivial safety constraints. Compared to several model-free baselines, our results show a significant reduction of training samples and total incurred cost in producing adequate control policies.

\section{RELATED WORK}
A large body of work considers safety by relieving the assumptions of canonical RL (e.g., sim-to-real transfer learning, offline RL, or imitation learning). However, we believe there is universal merit in designing agents that can safely learn from and interact with their intended environment and focus on such approaches. 

% Such avenues may involve \textit{sim-to-real} transfer learning\ \cite{andrychowiczLearningDexterousInhand2020}, offline RL\ \cite{urpiRiskAverseOfflineReinforcement2020}, or imitation learning and effectively dodge the risk of real-world exploration by lifting the assumption of direct interaction with the target domain. We believe, there is a universal merit in designing agents that can safely learn from and interact with their intended environment (possibly in combination with the above-mentioned techniques) and will thus focus on these approaches. 

An overview of safe RL is provided by Garcia and Fernandez\ \cite{garciaComprehensiveSurveySafe2015}. One line of research leverages gradient or parameter projection methods\ \cite{uchibeConstrainedReinforcementLearning2007,yangProjectionBasedConstrainedPolicy2019} to project performance-oriented updates to a constraint-satisfactory set. Moldovan and Abbeel restrict the set of feasible solutions to ergodicity-maintaining policies, that is, the ability to return to any previously observed state\ \cite{moldovanabbeel2012}. Similarly, Perkins and Barto consider Lyapunov system stability and utilize domain knowledge to define safe base-controllers as an agent's action space\ \cite{perkinsLyapunovDesignSafe2002a}. Dalal et al. relieve the need for domain-specific base actions by constructing safety layers with a safe state-wise action correction\ \cite{dalal2018}. Chow et al. project risky actions onto a feasible safety layer, characterized by linearized Lyapunov constraints\ \cite{chowLyapunovbasedApproachSafe2018b}. In contrast, methods based on Lagrangian duality solve the dual problem for Lagrange multipliers to obtain an unconstrained lower-bound on the original objective\ \cite{chowRiskConstrainedReinforcementLearning2018}. Achiam et al. approximate changes in expected cost with the KL divergence between adjacent policies\ \cite{achiamConstrainedPolicyOptimization2017a}, providing a forward-view on constraint satisfaction during optimization.

Model-based approaches for safety have been considered in the context of both model-predictive control (MPC)\ \cite{aswaniProvablySafeRobust2013,kabzanLearningBasedModelPredictive2019} and safe RL. Koller et al. use Gaussian Processes (GP) to predict dynamics with state-dependent uncertainty estimates and construct an MPC that guarantees trajectories returnable to safe state regions\cite{Koller2019Learningbased}. Berkenkamp et al. extend the use of GPs by incrementally exploring safe state regions under assumptions of Lipschitz continuity\ \cite{Berkenkamp2017}. As a more scalable approach to mitigating model-bias, Kurutach et al. integrate deep ensembles\ \cite{Lakshminarayanan2017} with trust-region policy optimization (TRPO). While Kurutach et al. restrict the task horizon to maintain feasible prediction horizons, Janner et al. propose a branched rollout scheme where short model- trajectories are started from previously observed off-policy states\ \cite{jannerWhenTrustYour2019a}. 
% Despite their expressiveness and capability of providing uncertainty estimates, non-parametric Bayesian approaches such as GPs suffer from limited scalability. To this end, Gal and Ghahramani, and Lakshminarayanan et al. posit that neural networks (NN) can approximate Bayesian uncertainty estimates by applying dropout or ensemble bootstrapping\ \cite{galDropoutBayesianApproximation2016a,Lakshminarayanan2017}. Leveraging an ensemble of NNs to mitigate maleffects of model bias, Kurutach et al. integrate trust-region policy optimization into the MBRL setting to improve sample efficiency\ \cite{kurutachModelEnsembleTrustRegionPolicy2018}. Notably, Kurutach et al. limit the task horizon to a length tractable under accumulating model-prediction errors. On this account, Janner et al. propose a branched rollout scheme where short on-policy model-generated trajectories are started from off-policy states stored in a replay buffer\ \cite{jannerWhenTrustYour2019a}. 
\section{PRELIMINARIES}
Throughout this work, a finite Constrained Markov Decision Process (CMDP)\ \cite{altmanConstrainedMarkovDecision1999} of the tuple $(\mathcal{S}, \mathcal{A}, r, C, P, \mu, D)$ serves as the default problem framework, where $\mathcal{S}$ is the state space, $\mathcal{A}$ is the action space, $r:S\times A \times S \rightarrow \mathbb{R}$ is the reward function, $C$ is the set of safety-associated cost functions ${\{c_i: \mathcal{S} \times \mathcal{A} \times \mathcal{S} \rightarrow \mathbb{R}} \}$, ${P: \mathcal{S} \times \mathcal{A} \times \mathcal{S} \rightarrow [0,1]}$ is the state transition probability, $\mu: \mathcal{S} \rightarrow [0,1]$ is the start state distribution, and $D$ is the set of user-defined cost limits $\{d_i \in \mathbb{R}\}$ associated with $c_i$. Throughout this work, we denote dependency on the cost functions with a subscript $c_i$. The expected discounted return and cost return of a stochastic policy $\pi(a|s)$ over actions $a \in \mathcal{A}$ can accordingly be expressed as $J(\pi)= \E_{\tau \sim \pi}[\sum_{t=0}^\infty \gamma^t r(s_t, a_t, s_{t+1})]$ and $J_{c_i}(\pi)= \E_{\tau \sim \pi}[\sum_{t=0}^\infty \gamma^t c_i(s_t, a_t, s_{t+1})]$. Trajectories $\tau=\{s_0, a_0, s_1,... \}$ with a start state $s_0$ are produced by drawing actions from the policy $a \sim \pi(a|s)$, and successor states from the true transition probability $s' \sim P(s'|s,a)$. In this setting, we consider policies that satisfy the constraint $J_{c_i}(\pi) \leq d_i, \, \forall i$ to be safe, such that a set $\Pi_c$ of feasible policies can be defined as $ \Pi_{c} = {\pi \in \Pi : \forall i, J_{c_i}(\pi)\leq d_i}$. The objective for constrained policy search can thus be described as the maximization of expected returns, subject to constraint satisfaction according to
\begin{equation}
\begin{aligned}
    & \text{maximize}_\pi \, \, J(\pi), \\
    & \text{subject to} \, \, J_{c_i}(\pi) \leq d_i, \, \forall i
\end{aligned}    
\end{equation}
The expected value of discounted trajectory returns $R(\tau) = \sum_{t=0}^\infty \gamma^t r(s_t,a_t,s_{t+1})$, conditioned on a start state or an initial state-action tuple, are defined as the value function $V^\pi(s)$ and the state-action value $Q^\pi(s,a)$. 
The difference between these identities is referred to as the advantage function $A^\pi(s,a)$ such that
\begin{align} \label{v} 
    V^\pi(s) &= \E_{\tau \sim (\pi,P)} \big[ R(\tau|s_0=s) \big], \\\label{q}
    Q^\pi(s, a) &= \E_{\tau \sim (\pi,P)} \big[ R(\tau|s_0=s, a_0=a) \big], \\\label{adv}
    A^\pi(s, a) &= Q^\pi(s,a) - V^\pi(s).
\end{align}
In the CMDP setting, we express cost-related counterparts to the functions (\ref{v}),(\ref{q}), and (\ref{adv}) as $V^\pi_{c_i}(s)$, $Q^\pi_{c_i}(s, a)$, $A^\pi_{c_i}(s, a)$. We now consider a learned model transition probability $P_m(s'|s, a) \neq P(s'| s, a)$ for which we indicate dependency with a subscript $m$ and accordingly write $V_m^\pi(s)$, $Q_m^\pi(s, a)$,  and $A_m^\pi(s, a)$ for value- and advantage functions under model dynamics. Often, it is useful to express expectations over the stationary discounted state distribution $d^\pi(s)$, which summarizes the discounted probability of visiting state $s$ given a policy $\pi = \pi(a|s)$ and a transition probability $P=P(s'|s,a)$. Kakade and Langford\ \cite{kakadeApproximatelyOptimalApproximate2002a} accordingly express the difference in expected returns between arbitrary policies $\pi$ and $\pi'$ as 
\begin{align} \label{kakade_diff}
    \Delta J(\pi, \pi') = J(\pi') - J(\pi) = \frac{1}{1-\gamma} \E_{\substack{s \sim d^{\pi'} \\ a \sim \pi'}} [A^\pi(s,a)] \,.
\end{align}
While this identity seemingly makes a natural candidate objective for return maximization, the above expression renders intractable for sampling-based optimization since sampling, and accordingly, the estimation of expected values can typically only take place under a fixed policy. To this end, Schulman et al.\ \cite{schulmanTrustRegionPolicy2015} and Achiam et al.\ \cite{achiamConstrainedPolicyOptimization2017a} refine identity (\ref{kakade_diff}) for a lower-bound on the return difference between policies given by
\begin{align} \label{cpobound}
    \Delta J(\pi, \pi') \geq \frac{1}{1-\gamma} \E_{\substack{s \sim d^{\pi} \\ a \sim \pi'}} \bigg[ A^\pi(s,a) - \frac{2\gamma \delta}{1-\gamma} D_{TV} (\pi'\|\pi)[s]\bigg] \,,
\end{align}
where $\delta = \text{max}_s|\E_{a \sim \pi'}[A^\pi(s,a)]|$ and $D_{TV}(\pi'\|\pi)[s]$ is the total variation distance between $\pi'$ and $\pi$. Notably, expression (\ref{cpobound}) considers expectations over the state distribution $d^\pi(s)$, thus providing a tractable objective for sample-based optimization with a fixed policy $\pi$.

\section{CONSTRAINED POLICY OPTIMIZATION UNDER MODEL UNCERTAINTY}
\subsection{Performance Boundaries Under Model Dynamics}
We now describe how performance differences between policies with respect to expected returns $\Delta J(\pi, \pi')$ and expected cost returns $\Delta J_c(\pi, \pi')$ can be connected to a model-based setting. In particular, we are interested in how real-world performance relates to an optimization objective that assumes access only to distributions under the model dynamics $P_m(s'|s,a)$. 
% In order to guarantee real-world safety and performance improvement, it is of natural interest how the mismatch of optimizing over $P_m(s'|s,a)$ rather than $P(s'|s,a)$, and $A^\pi_m(s,a)$ rather than $A^\pi(s,a)$ affects policy performance. 
To this end, we derive Boundary \ref{boundary1}, which relates the \textit{relative performance term} $L^{\pi}_m(\pi')$ to an error term induced by this mismatch. We label $ \displaystyle L^{\pi}_m(\pi')=  \E_{\substack{s \sim d_{m}^{\pi} \\ a \sim \pi}} [( \frac{\pi'(a|s)}{\pi(a|s)}) A^{\pi}_m(s,a)]$ a relative performance term as it connects policy $\pi'$ to a fixed policy $\pi$ through the likelihood-ratio of advantageous actions over model states, making it a suitable candidate for optimization.
\begin{manualtheorem}{4.1}\label{boundary1}
Let $\Delta J(\pi, \pi')$ be the true difference in expected returns $J(\pi') - J(\pi)$ between arbitrary policies $\pi'(a|s)$ and $\pi(a|s)$. Then $\Delta J(\pi', \pi)$ can be bounded by the following expression under model dynamics $P_m(s'|s,a)$: \\
\begin{equation}
\begin{aligned} 
    & \Delta J(\pi, \pi')
    \lesseqgtr \frac{L^{\pi}_m(\pi')}{1-\gamma}
        \pm \frac{4 \delta^{max} \epsilon}{1-\gamma} \\
& \text{where} \\
    & L^{\pi}_m(\pi')
    =  \E_{\substack{s \sim d_{m}^{\pi} \\ a \sim \pi}}
        \big[
            \big( \frac{\pi'(a|s)}{\pi(a|s)}\big) A^{\pi}_m(s,a)
        \big]\\
    & \delta^{max} = \max_{s,a,s'} \big| r(s,a,s') + V^{\pi}_{m}(s') - V^{\pi}_{m}(s) \big| \\
    & \epsilon = \epsilon_{\pi}\epsilon_m
        + \frac{\gamma}{1-\gamma}
        (
            \epsilon_{\pi}^2
            + 2 \epsilon_{\pi} \epsilon_{m}
        )\\
    & \epsilon_{\pi}
    = \max_{s}D_{TV}(\pi'(\cdot|s)||\pi(\cdot|s)) \\
    & \epsilon_{m}
    = \max_{s,a}D_{TV} \big(P(\cdot|s,a)||P_m(\cdot|s,a)\big)
\end{aligned}    
\end{equation}
The error terms $\epsilon_\pi$ and $\epsilon_m$ refer to the maximum total variation distance between policies $D_{TV}(\pi'||\pi) = \frac{1}{2} \sum_a |\pi(a|s)-\pi(a|s)|$ and state transition probabilities $D_{TV}(P||P_m) = \sum_{s'} |P'(s'|s,a)-P_m(s'|s,a)|$. The model state-value function $V_m^\pi(s)$ and the model advantage function $A^\pi_m(s,a)$ may be replaced by their cost-associated counterparts. $a \lesseqgtr b \pm c$ is read as $b+c \geq a \geq b - c$. \\
Proof. See supplementary material A.
\end{manualtheorem}
\paragraph{Implications of Boundary \ref{boundary1}}
Intuitively, Boundary \ref{boundary1} quantifies an error term $\epsilon$ that arises from expressing $\Delta J(\pi, \pi')$ over a state-distribution $d^\pi_m$ that is shifted by a disparate transition probability $P_m$ and policy $\pi$. The penalizing term comprises three sources of error, namely
\begin{enumerate}
    \item policy divergence between $\pi(a|s)$ and $\pi'(a|s)$ (in $\epsilon_\pi$),
    \item divergence between the state transition probabilities $P(s'|s,a)$ and $P_m(s'|s,a)$ (in $\epsilon_m$),
    \item the maximum temporal difference residual $\delta^{\text{max}}$ of model values $V^{\pi}_m$.
\end{enumerate}
Interpreting Boundary \ref{boundary1} for constraint satisfaction allows us to similarly relate the expected change in cost returns to the relative (cost) performance term $ \displaystyle L^{\pi}_{m,c}(\pi')=  \E_{\substack{s \sim d_{m}^{\pi} \\ a \sim \pi}} [( \frac{\pi'(a|s)}{\pi(a|s)}) A^{\pi}_{m,c}(s,a)]$. Given a safe policy $\pi$, the safety of a candidate policy $\pi'$ is thus guaranteed when
\begin{equation}
\begin{aligned}
    J_{c}(\pi') 
    &= J_{c}(\pi) + \Delta J_{c}(\pi', \pi) \\
    &\leq J_{c}(\pi) +
    \frac{L^{\pi}_{m, c}(\pi')}{1-\gamma}
        + \frac{4 \delta_{c}^{\text{max}} \epsilon}{1-\gamma}
        \overset{!}{\leq} d_{c}.
\end{aligned}    
\end{equation}
is satisfied. The boundary's tightness is determined by the penalizing right-hand side and motivates keeping the error terms $\epsilon_m$, $\epsilon_\pi$, and $\delta^{\text{max}}$ small during optimization. The boundary is tight for $\epsilon_\pi = \epsilon_m = 0$, as $L^{\pi}_{m}(\pi')$ and the penalty term vanish with the definition of the advantage function $\E_{a \sim \pi} [A_{\pi}^m(s,a)] = 0$. Notably, the penalty term vanishes for equal policies $\epsilon_\pi = 0$, thus supporting the intuitive notion that the same policy yields the same results in the true CMDP regardless of model errors.

% \paragraph{Relations to Previous Work}
% The boundaries are tight with respect to the errors $\epsilon_\pi$ and $\epsilon_m$, as $L^{\pi}_m(\pi')$ and the penalty term vanish with the definition of the advantage function $\E_{a \sim \pi} [A_{\pi}^m(s,a)] = 0$ with $\epsilon_\pi = \epsilon_m = 0$. When assuming an equality $V^{\pi}_m = V_{\pi}$ for a perfect model $\epsilon_m = 0$, the boundaries correspond to Theorem 1 by Schulman et al.\ \cite{schulmanTrustRegionPolicy2015} and Corollary 3.6 by Pirotta et al.\ \cite{pirottaSafePolicyIteration2013}. In comparison to Achiam et al.\ \cite{achiamConstrainedPolicyOptimization2017a}, our boundary comprises an additional factor of $\text{max}_{s'}$ in the maximum TD residual $\delta^{\text{max}}$, which originates from upper bounding the mismatch in one-step transitions between $P(s'|s,a)$ and $P_m(s'|s,a)$. In contrast to Theorem 4.1 by Janner et al.\ \cite{jannerWhenTrustYour2019a}, the presented boundaries pertain to return differences between distinct policies under the true dynamics rather than comparing returns under model-dynamics with true returns.
% This is further illustrated by the fact that for equal policies $\epsilon_\pi = 0$,
% \begin{align*}
%     \Delta J(\pi, \pi') &\lesseqgtr 
%     \frac{L^{\pi}_m(\pi)}{1-\gamma} \pm 0 = \frac{1}{1-\gamma} \E_{\substack{s \sim d_m^{\pi} \\ a \sim \pi}} \big[ A^{\pi}_m(s,a)\big] = 0,
% \end{align*}
% supporting the intuitive notion that the same policy yields the same results in the true CMDP, regardless of model errors. 

\subsection{Practical Algorithm}
Inspired by these theoretical results, we aim to derive a practical algorithm that reconciles model errors with approximate safety guarantees. Recall that boundary\ \ref{boundary1} motivates small model-errors $\epsilon_m$ and policy-divergences $\epsilon_\pi$ for guaranteed constraint satisfaction. In practice, literal obedience to this notion is intractable for several reasons: 1.) the maximum terms over the state space $\mathcal{S}$ and the action space $\mathcal{A}$ are intractable to compute and may pose a prohibitively heavy penalty on policy updates. 2.) While policy divergence around a fixed policy $\pi$ is controllable during optimization, our influence on the model-error $\epsilon_m$ is limited and initially unknown. To circumvent these limitations, we modify boundary \ref{boundary1} by replacing the maximum terms in $\epsilon_\pi$ and $\epsilon_m$ with their expected values over $d^\pi_m$ and upper bound the total variation distance with the KL divergence through Pinsker's inequality: $D_{TV}(p\|q) \leq \sqrt{D_{KL}(p\|q)/2}$. Moreover, instead of limiting the joint penalty term $\epsilon$, we restrict $\epsilon_\pi$ and $\epsilon_m$ individually to a maximum value, yielding a 
% Furthermore, we drop the undiscounted product $\epsilon_m \epsilon_\pi$ with $\gamma/(1-\gamma) \gg 1$ for $\gamma$ close to $1$, thus yielding the new penalty term
% \begin{align} \label{penalty}
%     & \bar{\epsilon} = \\ \nonumber
%     & \E_{\substack{s \sim d^\pi_m \\ a \sim \pi}} \Big[D_{KL}(\pi'\|\pi)][s] + 2 \sqrt{D_{KL}(\pi'\|\pi)[s]D_{KL}(P\|P_m)} \Big].
% \end{align}
% \paragraph{Policy Optimization}
% Based on expression\ \ref{penalty}, we adopt 
a constrained optimization problem that closely resembles trust-region methods as formulated by Schulman et al.\ \cite{schulmanTrustRegionPolicy2015} and Achiam et al.\ \cite{achiamConstrainedPolicyOptimization2017a}. The problem is then given by
\begin{equation}\label{problem}
\begin{aligned}
    \text{maximize}_{\pi'}& \E_{\substack{s \sim d_{m}^{\pi} \\ a \sim \pi}}
        L^{\pi}_m(\pi') \\ 
    \text{subject to}   & \E_{s \sim d_{m}^\pi} [D_{KL}(\pi'\|\pi)] \leq d_\pi \\ 
                        & \E_{\substack{s \sim d_{m}^\pi \\ a \sim \pi}} 
                        [   J_{c}(\pi) + 
                            \frac{L^{\pi}_{m, c}(\pi')}{1-\gamma}] \leq d_{c} \\ 
                        & \E_{\substack{s \sim d_{m}^\pi \\ a \sim \pi}} [D_{KL}(P\|P_m)] \leq d_m .
\end{aligned}    
\end{equation}
\paragraph{Policy Optimization}
% By restricting the penalizing components of $\bar{\epsilon}$ to limits $d_\pi$ and $d_m$, we aim to preserve approximate guarantees on policy improvement and constraint satisfaction. While this expression deviates from a literal interpretation of equation \ref{penalty}, this formulation allows us to adopt the constrained policy optimization (CPO) algorithm described by Achiam et al.\ \cite{achiamConstrainedPolicyOptimization2017a}. 
Framing the model-based policy optimization problem as (\ref{problem}) allows us to straightforwardly integrate Constrained Policy Optimization (CPO)\ \cite{achiamConstrainedPolicyOptimization2017a} as a policy search algorithm. The relative performance terms $L^{\pi}_{m}(\pi')$ and $L^{\pi}_{m, c}(\pi')$ are composed of a stochastic policy $\pi_\theta(a|s)$ and separate value functions $V_\eta (s)$, $V_{c,\zeta}(s)$ parametrized with feedforward NNs. The constrained problem\ (\ref{problem}) can then be approximated by linearizing the objective and cost constraints around $\theta$ while employing a second-order Taylor expansion around $\theta$ for policy divergence. CPO uses Lagrangian duality to solve the approximation to the primal problem by determining Lagrange multipliers through an analytical solution to the dual. Policy and value parameters are updated alternatingly in an actor-critic fashion.

\paragraph{Model Learning and Usage}
Before proceeding to address how the constraint on model errors in\ (\ref{problem}) can be approximated and enforced, we describe our dynamics model architecture and usage. We generate model trajectories from recursive next state predictions by employing an $M$-sized ensemble $E$ of feedforward NNs with parameters $\vartheta$. The outputs of the ensemble parametrize a multivariate normal distribution with the mean $\bm{\mu}_{\vartheta m} = \bm{\mu}_{\vartheta m}(\bm{s},\bm{a})$ and a diagonal covariance matrix $\bm{\Sigma}_{\vartheta m} = \bm{\Sigma}_{\vartheta m}(\bm{s},\bm{a})$ with entries $\bm{\sigma}^2_{\vartheta m} = \bm{\sigma}^2_{\vartheta m}(\bm{s},\bm{a})$. The state transition probabilities predicted by the ensemble thus become $P_{\vartheta E} = \{ P_{\vartheta m}(\bm{s}'|\bm{s},\bm{a})= \mathcal{N}(\bm{\mu}_{\vartheta m}, \bm{\Sigma}_{\vartheta m}) \}$, $m\in [1,...,M]$. As Chua et al.\ \cite{chuaDeepReinforcementLearning2018a} point out, variance predictions of probabilistic networks can lead to erratic values for out-of-distribution points. Instead of minimizing the negative likelihood, we resort to an altered objective that we find to behave more smoothly. We minimize the loss function 
\begin{align}
    \mathcal{L}_m = \frac{1}{N} \sum_{n} \| \bm{\mu}_{\vartheta m,n} - \Delta \bm{s_{n}} \|^2_2 + \| \bm{\sigma}^2_{\vartheta m} - \text{SE}_{n} \|^2_2 \,.
\end{align}
The mean-prediction targets $\Delta \bm{s_n} = \bm{s'_n} - \bm{s_n}$ are the state-changes observed between subsequent state transitions, and $\text{SE}_\mu$ describes the squared error $\| \bm{\mu}_{\vartheta m,n} - \Delta \bm{s_n} \|^2_2$ for which we stop gradient propagation during training. For every state prediction, our model additionally produces a reward estimate $\hat{r}_\vartheta(\bm{s}, \bm{a})$ using the same loss-function. Cost- and termination-functions are computed deterministically as $c(\bm{s}, \bm{a})$ and $T(\bm{s}, \bm{a})$ to prevent further bias of expected cost-returns. All parameters $\vartheta$ are initialized randomly, and batch orders are shuffled independently for each model in the ensemble to facilitate independent predictions. For inference, predictions are computed recursively by $\hat{\bm{s}}_{t+1} = \bm{s_t} + \Delta \hat{\bm{s}}_t$, $\Delta \hat{\bm{s}}_{t} = \bm{\mu}_{\vartheta m}(\bm{s},\bm{a})$ with a randomly chosen model index $m \sim \mathcal{U}(1,M)$. Note that we do not sample from the parametrized normal distribution $P_{\vartheta m}(\bm{s}'|\bm{s},\bm{a})$, which we find behaves more stable when the assumption of normal-distributed targets does not hold. The model-generated trajectories follow the branched rollout scheme described by Janner et al.\ \cite{jannerWhenTrustYour2019a}, where starting states are sampled from an off-policy buffer, thus allowing coverage of the whole task horizon even for short rollout horizons. It is worth noting that this procedure shifts the model state distribution $d^\pi_m$ towards a joint distribution over past policies, causing a deviation from the theoretically derived constraints. To mitigate this effect, we sample starting states from trajectories $\tau_{\pi_k}$ sampled under past policies $\pi_k$ according to a Boltzmann distribution $Pr(\tau_{\pi_k})= \eta \exp(-\beta \E_{s,a \sim \tau_{\pi k}}[D_{KL}(\pi\|\pi_k)])$ where $\eta$ is a normalizing constant and $\beta$ a temperature parameter that regulates the preference for samples collected under similar policies.
\paragraph{Predictive Model Uncertainty Estimation} 
We now aim to approximate the KL divergence between predicted dynamics and the true transition probability $\E_{\substack{s \sim d_{m}^\pi \\ a \sim \pi}} [D_{KL}(P\|P_m)]$ using a scale-invariant, state-dependent measure on ensemble disagreement. Lacking knowledge of the true dynamics, we invoke the average KL divergence between transition distributions within the ensemble $E$ as a surrogate approximation on true model-errors. For the normally distributed predictions $P_{\vartheta E}$, we can compute this identity in closed form and approximate $D_{KL}(P\|P_m) \approx D_{KL}^E(\bm{s}, \bm{a})$ by
\begin{align}
    D_{KL}^E(\bm{s}, \bm{a}) = \frac{1}{M(M-1)} \sum_m^M \sum_n^{M} D_{KL}(P_{\vartheta m} \| P_{\vartheta n}),
\end{align}
where $P_{\vartheta}$ is short for $P_{\vartheta}(\bm{s},\bm{a})$. Accordingly, we denote the expected model-uncertainty over model-generated samples as $\bar{D}_{KL}^E(\pi) = \E_{s \sim d^{\pi}_m, a \sim \pi}[D_{KL}^E(s, a)]$.
\paragraph{Adaptive Resampling}\label{resampling}
Having established a measure for predictive model error estimation without reevaluating real samples, we now consider how constraints on maximum expected uncertainty can be enforced. A straightforward consequence of (\ref{problem}) is to minimize the expected KL divergence between predictions and ground-truth dynamics, which reduces to maximum likelihood estimation for empirical distributions. Still, the capacity for model improvement may be limited, or policy updates may lead to frequent encounters with unseen state-action subspaces. We thus devise an adaptive resampling scheme in which average model uncertainty is reduced through sample mixing. Diluting model-generated samples with real data allows us to freely enforce a constraint on average model-uncertainty by adjusting a mixing factor $\alpha(\pi)$ such that
\begin{equation} \label{alpha}
\begin{aligned}
    \big(1-\alpha(\pi)\big)\bar{D}_{KL}^E(\pi) \leq d_m
\end{aligned}    
\end{equation}
is satisfied. Intuitively, $\alpha(\pi)$ is increased when estimated model-errors are large, either through rapid policy change or deteriorating model-performance (e.g., through overfitting) and prevents undue reliance on model-generated trajectories when safe policy improvement in the true CMDP can not be guaranteed. 

\paragraph{Adaptive Rollout Horizons}
Lastly, we describe a procedure to contain accumulating model errors in trajectories built on recursive one-step state predictions. Note that we have so far expressed expected model-errors over the stationary model state distribution $d^\pi_m$, providing little guidance on trajectory-wise error development. To this end, we propose limiting the accumulated model-uncertainty $\sum_{t=0}^h D_{KL}^E(s_t, a_t)$ along model trajectories of variable horizons $h$. Denoting a limit on cumulative KL divergence $d_H$, we obtain model-generated trajectories with a termination condition according to
\begin{equation}\label{roll}
\begin{aligned}
    \tau_m^\pi (s_0) &= \{s_0, a_0, \hat{s}_{t+1}, a_{t+1}, ..., \hat{s}_{h}, a_{h}\} \\
    \text{s.t.} &\, \sum_{t=0}^{h} D_{KL}^E(s_t, a_t) \leq d_H \, .
\end{aligned}
\end{equation}
A state-dependent adaptation of rollout horizons in this form bears two practical benefits: 1) It allows for extended rollouts when predictions are reliable, thus reducing policy-induced distribution shifts; 2) It naturally integrates with the constraint (\ref{alpha}), as lower per-step uncertainty increases the maximum rollout horizon; 

We acknowledge that the constraint limits on average uncertainty $d_m$ and cumulative uncertainty $d_H$ can be unintuitive to determine in practice. Rather than explicitly setting $d_H$ and $d_m$, we suggest performing an initial model training on a small set of samples collected under an untrained policy $\pi_0$. The limit $d_m$ is then determined by measuring the average model uncertainty $\bar{D}_{KL}^{E}(\pi_0)$ on the initial dataset and defining $d_m = \big(1-\alpha(\pi_0)\big)\bar{D}_{KL}^{E}(\pi_0)$ for an initial mixing ratio $\alpha_0$. Similarly, the constraint on cumulative uncertainty is set to $d_H=\sum_{t=0}^{h_0} D_{KL}^E(s_t, a_t)$ for an initial rollout horizon $h_0$.  This procedure allows us to define the constraint values\ \ref{alpha} and\ \ref{roll} through more intuitive parameters on an initial mixing ratio and horizon while relating model-uncertainties during training to early model performance.

\section{EVALUATION ON CONSTRAINED ROBOT LOCOMOTION}
\subsection{Experiment Setup}
\begin{figure}
    \begin{center}
    \begin{tabular}{cccc}
    \subcaptionbox{AntSafe \label{antsafe}}{\includegraphics[width = 1.5in]{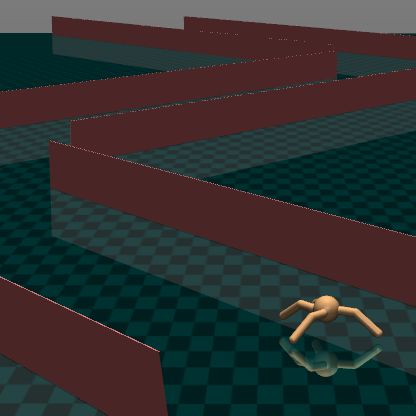}} &
    \subcaptionbox{HalfCheetahSafe \label{hcsafe}}{\includegraphics[width = 1.5in]{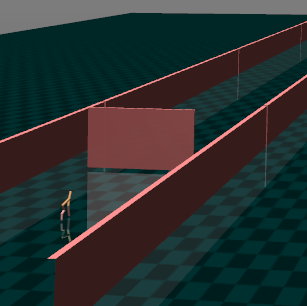}} \\
    \subcaptionbox{Ant robot \label{ant}}{\includegraphics[width = 1.5in]{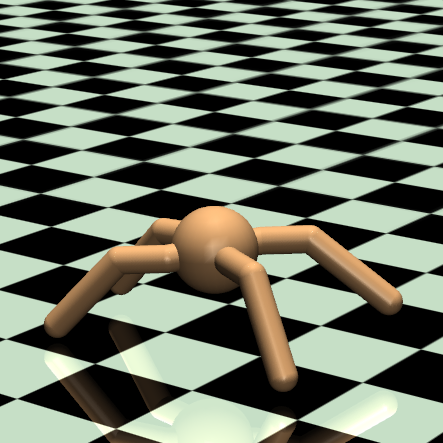}} & 
    \subcaptionbox{Circle task \label{circle}}{\includegraphics[width = 1.5in]{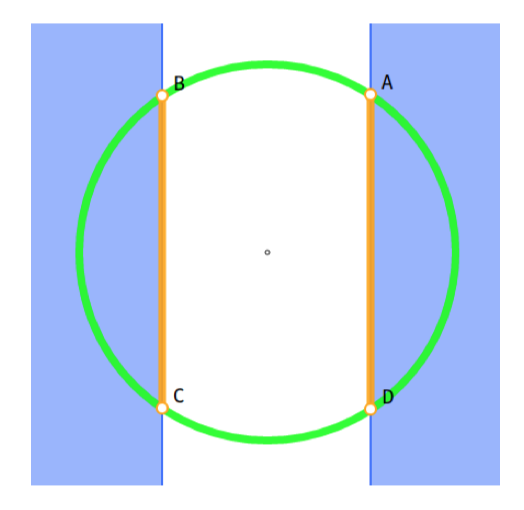}} 
    \end{tabular}
    \end{center}
    \caption{(a): The AntSafe environment with overthrowable walls. (b): The HalfCheetahSafe environment with a moving heading obstacle. (c): A close-up of the quadruped Ant robot. (d): Layout of the circle task. Rewards are maximized along the green circle while entering the blue region is constrained\ \cite{achiamConstrainedPolicyOptimization2017a}. }
\end{figure}

We evaluate the efficacy of our algorithm, labeled Constrained Model-Based Policy Optimization (CMBPO), on several simulated high-dimensional robot locomotion tasks with continuous state- and action spaces. Our aim is to better understand the practical merits and limitations of the proposed approach by assessing the results in light of the following questions:
\begin{enumerate}
    \item Can CMBPO maintain safety constraints throughout training in high-dimensional state- and action-spaces?
    \item Does the usage of model-generated samples improve the sample efficiency of constraint policy search?
    \item How do the proposed techniques affect asymptotic performances and how do these compare to baseline algorithms?
    \item How do sample mixing and adaptive rollout horizons affect policy performance and constraint satisfaction?
\end{enumerate}
In our experiments, we consider a quadruped 8-DoF \textit{Ant} robot (see Fig. \ref{antsafe} \& \ref{ant}) and a 2-dimensional 6-DoF \textit{HalfCheetah} robot (see Fig. \ref{hcsafe}) in the Mujoco\ \cite{todorovMuJoCoPhysicsEngine2012} physics simulator. Both robots are controlled through continuous time-discrete torque signals to the actuators and receive positions, orientations, and linear \& angular velocities of all body joints, as well as the global center of mass position as observations.
We perform an evaluation on three distinct tasks, which demand a trade-off between greedy return maximization and safety constraint satisfaction: (1) In the AntSafe task (see Fig.\ \ref{antsafe}), the agent receives rewards for its traveling-velocity in x-direction while constrained to maintain a safe distance to overthrowable walls. (2) The HalfCheetahSafe task (see Fig.\ \ref{hcsafe}) adopts the same reward-function as AntSafe but constrains unsafe proximity to a moving object in front of the agent. (3) The AntCircle task (see Fig.\ \ref{circle},\ \cite{achiamConstrainedPolicyOptimization2017a}) rewards an agent for running along a circle with the constraint of staying within a safe corridor.
\paragraph{Comparative Evaluation}
We compare CMBPO to three safe exploration algorithms, namely Constrained Policy Optimization (CPO)\ \cite{achiamConstrainedPolicyOptimization2017a}, Lagrangian Trust Region Policy Optimization (TRPO-L), Lagrangian Proximal Policy Optimization (PPO-L)\ \cite{rayBenchmarkingSafeExploration}, and an unconstrained version of TRPO\ \cite{schulmanTrustRegionPolicy2015}. 
\begin{figure*}[h]
    \begin{center}
        \resizebox{\textwidth}{!}{
        \input{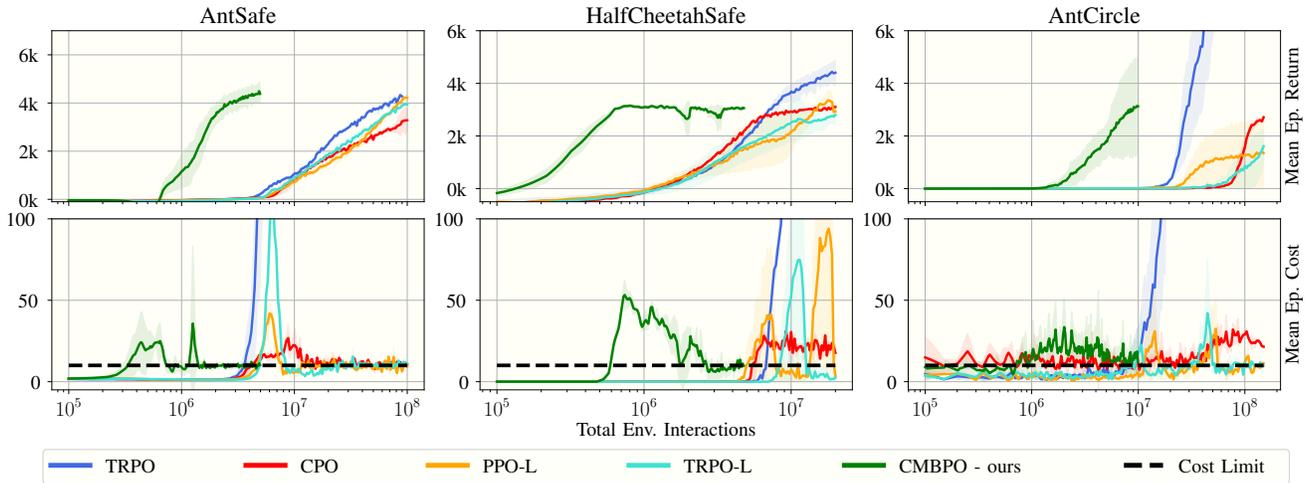}
        }
    \end{center}
    \caption{Training curves of CMBPO and 4 baselines. Solid lines correspond to mean curves of 4 random seeds for CMBPO and 2 seeds for baselines. Shaded regions represent $\pm1$ standard deviation across seeds. Curves are plotted against total environment interactions on a logarithmic scale, the constraint limit is highlighted in dashed black.}
    \label{exp}
\end{figure*}
The results of our experiment are shown in Fig.\ \ref{exp}. We observe that the unconstrained algorithm TRPO exceeds cost constraints on all experiments, highlighting the trade-off between greedy return maximization and constraint satisfaction in our environments. CMBPO reaches model-free asymptotic performances on all tested experiments with an increase in sample efficiency by 1-2 orders of magnitude. Compared to the safe baseline algorithms, CMBPO exhibits slightly less adherence to constant constraint satisfaction but succeeds in reaching safe policies asymptotically. We attribute this to lower-quality advantage estimates during early training phases, an error-source the model-generated trajectories are more susceptible to due to limited horizons.
\begin{table}[h]
\begin{center}
\begin{tabular}{|c||c|c|c|c|}
\hline
Env & CMBPO & CPO & PPO-L & TRPO-L\\
\hline
AntSafe@2500 & 18 & 527 & 422 & 385\\
HalfCheetahSafe@3000 & 58 & 100 & 515 & 223\\
AntCircle@2000 & 107 & 2395 & 1548 & 1576 \\
\hline
\end{tabular}
\end{center}
\caption{Cumulative Costs (in 1e3): Average cumulative costs over the entire training duration before producing a \textit{constraint-satisfactory} policy with given mean returns}
\label{cumcost}
\end{table}
However, due to the increased sample efficiency, cumulative costs (see Table \ref{cumcost}) received upon achieving well-performing policies are significantly reduced in comparison to the baseline algorithms. To our surprise, we find that CMBPO in some cases exceeds the asymptotic performance of model-free optimizers, which we suspect may be caused by exploration through temporary constraint violation or lower-variance gradients due to only performing expected state-transitions in model-trajectories.
\paragraph{Ablation on Sample Mixing and Adaptive Rollouts}
\begin{figure}[!h]
    \begin{center}
        \resizebox{.5\textwidth}{!}{
        \input{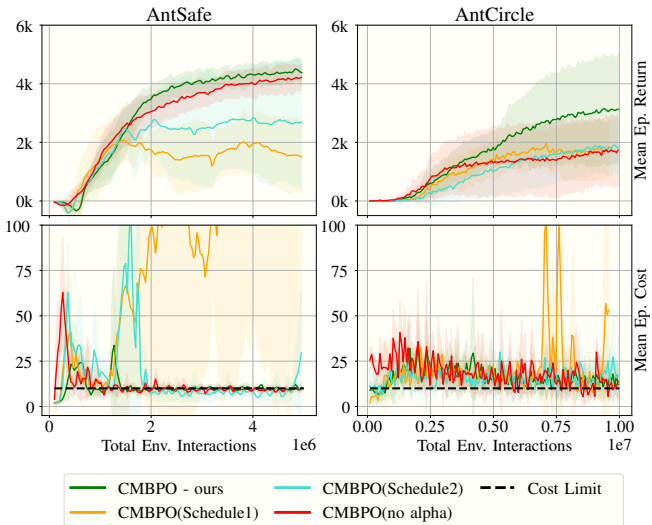}
        }
    \end{center}
    \caption{Ablation on sample mixing and adaptive rollouts: CMBPO(no alpha) optimizes with a constant real-to-model ratio of samples ($5\%$ on AntSafe and $3\%$ on AntCircle). CMBPO(Schedule1) increases rollout horizons from $3 \rightarrow 15$ over 1000 epochs. CMBPO(Schedule2) increases rollout horizons from $5 \rightarrow 25$ over 1000 epochs. All experiments performed on 4 random seeds.}
    \label{abl}
\end{figure}
Our ablation studies are aimed to better our understanding of the influence of sample mixing and adaptive rollouts on CMBPO's performance. Fig. \ \ref{abl} shows training curves for CMBPO compared to variants with a constant real-to-model sample ratio $\alpha$, and a rollout schedule with fixed horizons. For a fair comparison, both $\alpha$ and the rollout schedule were determined through a hyperparameter search. We observe that a combination of adaptive sample mixing and dynamic rollout horizons consistently outperforms the ablation baselines both in constraint satisfaction and final performance. The adaptive limitation of rollout horizons through cumulative model-uncertainty demonstrates a particularly significant difference to the scheduled horizon variant, which we found to frequently suffer from temporary or even permanent deterioration in value approximation quality. The stability of function approximation with bootstrapping is a well-known issue\ \cite{vanhasseltDeepReinforcementLearning2018} and highlights a dilemma of static rollout horizons in model-based RL: Short rollouts limit model-errors but increase the susceptibility to value instability due to bootstrapping, an issue exacerbated in the presence of potential contact dynamics as in AntSafe. 

In contrast, the influence of adaptive sample mixing appears less clear. We observe a slight tendency of the fixed-alpha variant towards local minimum convergence on AntCircle but remain sceptical due to the generally large performance variance on this environment. It is worth noticing, however, that fixed mixing ratios do not appear to learn earlier than the adaptive version despite training on fewer samples per epoch in early stages when model-inaccuracies are high. 

Lastly, we point out that preliminary experiments on the recently published benchmark suite safety-gym\ \cite{rayBenchmarkingSafeExploration} showed poor performances of CMBPO due to partial observability and strong covariances between state dimensions. We consider this the main limitation of our approach and regard more elaborate model architectures and a principled way of generating trajectories as important future avenues.
\addtolength{\textheight}{-1.5cm}   % This command serves to balance the column lengths
\section{CONCLUSION}
In this work, we have provided theoretical insights and a practical algorithm for model-based data augmentation of expectation-based constrained policy search algorithms. The derived approach is based on the realization that model-based RL faces a distinctive challenge when exact estimates of expected constraint violations are imperative. We believe this notion entails fundamental questions on how reliance on a learned model should be determined in the face of changing policies and cumulative prediction errors. On this account, we demonstrated that a combination of adaptive sample mixing and dynamic limitation of rollout horizons can provide assistance in maintaining stability and constraint satisfaction during training by preventing undue reliance on uncertain model predictions. We consider these ideas of particular importance to real-world applications in robotics where unstructured environments and changing agent behavior demand a flexible perspective on model-usage. We deem various model architectures with a focus on resolving temporal dependencies, such as attention-based, LSTM-based, or differential-equation-based methods, an important line of research to extend this approach's applicability to partially observable CMDPs. Furthermore, latent space models and policies may play an important role in using more generally applicable image-based observations. More dedicated usage of the model, for example by leveraging inherently safe model-interactions for exploration and information gain, may be a path to outperforming model-free approaches. Finally, a particularly exciting avenue is the application and evaluation of CMBPO for safe control on physical robots.

%%%%%%%%%%%%%%%%%%%%%%%%%%%%%%%%%%%%%%%%%%%%%%%%%%%%%%%%%%%%%%%%%%%%%%%%%%%%%%%%

%%%%%%%%%%%%%%%%%%%%%%%%%%%%%%%%%%%%%%%%%%%%%%%%%%%%%%%%%%%%%%%%%%%%%%%%%%%%%%%%

%%%%%%%%%%%%%%%%%%%%%%%%%%%%%%%%%%%%%%%%%%%%%%%%%%%%%%%%%%%%%%%%%%%%%%%%%%%%%%%%
%%%%%%%%%%%%%%%%%%%%%%%%%%%%%%%%%%%%%%%%%%%%%%%%%%%%%%%%%%%%%%%%%%%%%%%%%%%%%%%%

\bibliographystyle{IEEEtran}
\bibliography{IEEEabrv,root.bib}
\clearpage
\onecolumn
\section*{APPENDIX A}
\subsection{Proof of Performance Boundaries Under Model Uncertainty} \label{proof_boundary1}
The following proof is based on previous work by Achiam et al.\ \cite{achiamConstrainedPolicyOptimization2017a}, Schulman et al.\ \cite{schulmanTrustRegionPolicy2015}, and Kakade and Langford\ \cite{kakadeApproximatelyOptimalApproximate2002a}. The essence of the presented finding is that a relative performance boundary regarding returns under the true dynamics of the underlying CMDP is related to an optimization term and a penalizing error term under model transition probabilities $P_m(s'|s,a)$. To the best of our knowledge, the resulting boundaries are novel.
\subsubsection{Preliminaries}
The findings are defined for a finite CMDP as per earlier provided definitions. Identities, which refer to an item under the assumption of the learned state transition probability function $P_m(s'|s,a)$ are labeled with a subscript $m$. The start state distribution $\mu(s_0)$ is assumed to be equal both under the true MDP and under model usage. In this setting, the discounted stationary state distribution under an arbitrary policy $\pi(a|s)$ is defined as
\begin{equation} \label{d_pi'}
d^{\pi}(s) = (1-\gamma)\sum^{\infty}_{t=0}{\gamma^{t}Pr(S_{t}=s|\pi,P)} \, ,
\end{equation}
where $Pr(S_{t}=s|\pi)$ is the probability of visiting a state at timestep $t$. Similarly, the stationary model state distribution relates to the probability of visiting a state according to model dynamics as follows:
\begin{equation} \label{d_pi'_m}
d^{\pi}_{m}(s) = (1-\gamma)\sum^{\infty}_{t=0}{\gamma^{t}Pr(S_{t}=s|\pi,P_m)} \,.
\end{equation}
For later employed techniques, a representation of the above identities in the vector space is useful. For this, the following additional items are defined: \newline
The vectorized state visitation probability $\bm{p}_{\pi}^{t}(\bm{s}) \in \mathbb{R}^{|\mathcal{S}|}$ at time $t$ under true dynamics and model dynamics
\begin{align} \nonumber
    \bm{p}_{\pi}^{t} &= [Pr(S_t=s_0|\pi,P), ... , Pr(S_t=s_N|\pi,P)]^T, \\ 
    \bm{p}_{\pi,m}^{t} &= [Pr(S_t=s_0|\pi, P_m), ... , Pr(S_t=s_N|\pi, P_m)]^T.
\end{align}
The state transition matrix $\bm{P}^{\pi} \in \mathbb{R}^{|\mathcal{S}|\times|\mathcal{S}|}$ under true dynamics and model dynamics
\begin{align} \nonumber
    \bm{P}^{\pi}: P^{\pi}_{ij}  &= \sum_a P(s_j|s_i,a) \pi(a|s_i) \,, \\
    \bm{P}^{\pi}_m: P^{\pi}_{m, ij} &= \sum_a P_m(s_j|s_i,a) \pi(a|s_i) \,.
\end{align}
For a start state distribution vector $\bm{\mu} \in \mathbb{R}^{|\mathcal{S}|}$, the state visitation probabilities can thus be written as
\begin{align} \nonumber
    \bm{p}_{\pi}^{t} &= \bm{P}^{\pi} \bm{p}_{\pi}^{t-1}, &
    \bm{p}_{\pi,m}^{t} &= \bm{P}^{\pi}_m \bm{p}_{\pi,m}^{t-1}, \\
    &= (\bm{P}^{\pi})^t \bm{\mu}, &
    &= (\bm{P}^{\pi}_m)^t \bm{\mu},
\end{align}
such that
\begin{align}\label{state_dist} \nonumber
    \bm{d}^{\pi} & = (1-\gamma)\sum_{t=0}^{\infty}(\bm{P}^{\pi})^t \bm{\mu}, 
    &
    \bm{d}^{\pi}_m & = (1-\gamma)\sum_{t=0}^{\infty}(\bm{P}^{\pi}_m)^t \bm{\mu}, 
    \\
    & = (1-\gamma)(\bm{I}-\gamma \bm{P}^{\pi})^{-1} \bm{\mu},
    &
    & = (1-\gamma)(\bm{I}-\gamma \bm{P}^{\pi}_m)^{-1} \bm{\mu}.
\end{align}
By inserting the vectorized form $\bm{f} \in \mathbb{R}^{|\mathcal{S}|}$ of an arbitrary function $f:\mathcal{S}  \xrightarrow{} \mathbb{R}$, we obtain
\begin{align} \nonumber
    (\bm{I}-\gamma \bm{P}^{\pi})  \bm{d}^{\pi} & = (1-\gamma)\bm{\mu}, \\ \nonumber
    \Blangle (\bm{I}-\gamma \bm{P}^{\pi})  \bm{d}^{\pi} , \bm{f} \Brangle & =(1-\gamma) \Blangle \bm{\mu} , \bm{f} \Brangle,
    \\
    \E_{s\sim d^{\pi}}[f(s)]- \E_{\substack{s\sim d^{\pi} \\ a\sim \pi \\ s' \sim P}}[\gamma f(s')] &= (1-\gamma)\E_{s\sim \mu}[f(s)].
\end{align}
For $J(\pi)=1/(1-\gamma) \E_{s,a,s'\sim d^{\pi, a\sim\pi, s'\sim P}}[r(s,a,s')]$ and $J_m(\pi)=1/(1-\gamma) \E_{s\sim d^{\pi}_m, a\sim\pi, s'\sim P_m}[r(s,a,s')]$, the discounted returns of policy $\pi$ can be expressed for arbitrary functions $f(s)$ as
\begin{align} \label{J_pi'}
    J(\pi) &= \E_{s\sim \mu}[f(s)]+ \frac{1}{(1-\gamma)}\E_{\substack{s\sim d^{\pi} \\ a\sim \pi \\ s' \sim P}}[r(s,a,s')+\gamma f(s')-f(s)], \\ \label{J_pi'm}
    J_m(\pi) &= \E_{s\sim \mu}[f(s)]+ \frac{1}{(1-\gamma)}\E_{\substack{s\sim d^{\pi}_m \\ a\sim \pi \\ s' \sim P_m}}[r(s,a,s')+\gamma f(s')-f(s)].
\end{align}
(\ref{J_pi'm}) extends Lemma 1 by Achiam et al.\ \cite{achiamConstrainedPolicyOptimization2017a} for transition divergences. 
% For $f(s) = V^{\pi}(s)$, the right hand expected value in (\ref{J_pi'}) vanishes, as $J(\pi)$ is known to equal $\E_{s \sim \mu}[V^{\pi}(s)]$. We leave 
% Setting $f(s) = V^{\pi}(s)$ in (\ref{J_pi'm}), however, demonstrates a non-zero expected temporal difference residual $\E_{s \sim d_m^{\pi}, a\sim \pi, s'\sim P_m}[r(s,a,s')+\gamma f(s') -f(s)]$ since $\E_{s \sim \mu}[V^{\pi}(s)]$ does not fully account for the shifted $J_m(\pi)$.
Notably, Theorem 4.1 by Janner et al.\ \cite{jannerWhenTrustYour2019a} could be retrieved at this point by substracting\ (\ref{J_pi'm}) from\ (\ref{J_pi'}) such that a shift in policies and model-distribution between $J_m(\pi')$ and $J(\pi)$ is described. For this, the inner product space $\langle\cdot,\cdot\rangle:\mathbb{R}^{\mathcal{|S|}}\times\mathbb{R}^{\mathcal{|S|}}\xrightarrow\ \mathbb{R}$ is denoted $\langle\cdot,\cdot\rangle_s$ and \quad $\langle\cdot,\cdot\rangle:\mathbb{R}^{\mathcal{|S|}}\times\mathbb{R}^{\mathcal{|A|}}\times\mathbb{R}^{\mathcal{|S|}}\times\mathbb{R}^{\mathcal{|A|}}\xrightarrow\ \mathbb{R}^{\mathcal{|S|}}$ is denoted $\langle\cdot,\cdot\rangle_a$. 
% Expected values may be used interchangeably with their form in vector space (i.e., $\E_{a\sim \pi, s'\sim P}[f(s,a,s')]$ can represent a function of $s$ or a vector in $\mathbb{R}^{|\mathcal{S}|}$). 
The return difference between a policy $\pi'$ under model dynamics and a policy $\pi$ according to Janner et al.\ \cite{jannerWhenTrustYour2019a} is then described by
\begin{equation} \label{ret_diff_jjm}
\begin{split}
    J_m(\pi')-J(\pi) & = (1-\gamma) \big ( \Blangle \bm{d}^{\pi'}_m, \E_{\substack{a\sim \pi' \\ s'\sim P}}[\delta_f(s,a,s')]\Brangle_s - \Blangle \bm{d}^{\pi}, \E_{\substack{a\sim \pi \\ s'\sim P}}[\delta_f(s,a,s')]\Brangle_s \big )\\
        & = (1-\gamma) \big ( \Blangle \bm{d}^{\pi'}_m-\bm{d}^{\pi}, \E_{\substack{a\sim \pi' \\ s'\sim P}}[\delta_f(s,a,s')]\Brangle_s + \Blangle \bm{d}^{\pi}, \E_{\substack{a\sim \pi' \\ s'\sim P}}[\delta_f(s,a,s')]\Brangle_s - \Blangle \bm{d}^{\pi}, \E_{\substack{a\sim \pi \\ s'\sim P}}[\delta_f(s,a,s')]\Brangle_s \big )\\
        & = (1-\gamma)\big ( \Blangle \bm{d}^{\pi'}_m - \bm{d}^{\pi}, \E_{\substack{a\sim \pi' \\ s'\sim P}}[\delta_f(s,a,s')]\Brangle_s + \Blangle \bm{d}^{\pi}, \Blangle \bm{\pi'} - \bm{\pi}, \E_{s'\sim P}[\delta_f(s,a,s')]\Brangle_a \Brangle_s \big ).
\end{split}
\end{equation}
With Hölder's inequality (see also\ (\ref{holder})), the right hand side of (\ref{ret_diff_jjm}) can be related to an upper bound of the terms $$\text{max}_{s} D_{TV}(\pi'(a|s)\|\pi(a|s)) + 2 \, \text{max}_{s,a} D_{TV}(P_m(s'|s,a)\|P(s'|s,a))$$ and $$\text{max}_{s} D_{TV}(\pi'(a|s)\|\pi(a|s)),$$ resulting in Theorem 4.1 by Janner et al.\ \cite{jannerWhenTrustYour2019a}. Importantly, our following approach instead compares the true returns $J(\pi')$ and $J(\pi)$ directly and relates them to an optimization term that only assumes access to model dynamics.
\\
\\
\subsubsection{Proof}
Let $J(\pi)$ and $J(\pi')$ be the expected discounted returns of arbitrary policies $\pi'(a|s)$ and $\pi(a|s)$. Then the difference in true returns can be described by
\begin{align} \label{ret_diff}
    J(\pi')-J(\pi)=\frac{1}{1-\gamma} \big( \E_{\substack{s\sim d^{\pi'} \\ a\sim \pi' \\ s' \sim P}}[\delta_{f}(s,a,s')]-\E_{\substack{s\sim d^{\pi} \\ a\sim \pi \\ s' \sim P}}[\delta_{f}(s,a,s')] \big) \, .
\end{align}
Similarly to (\ref{ret_diff_jjm}), the terms used above can be related to the model state distribution $d_m^{\pi}$ by
\begin{align*}
\allowdisplaybreaks
    \E_{\substack{s\sim d^{\pi'} \\ a\sim \pi' \\ s' \sim P}}[\delta_{f}(s,a,s')]
    &= \Blangle \bm{d}^{\pi'} - \bm{d}_m^{\pi}, \E_{\substack{a\sim \pi' \\ s'\sim P}}[\delta_f(s,a,s')]\Brangle_s + \Blangle \bm{d}_m^{\pi}, \E_{\substack{a\sim \pi' \\ s'\sim P}}[\delta_f(s,a,s')] \Brangle_s \\ \nonumber
    &=
        \Blangle \bm{d}^{\pi'} - \bm{d}_m^{\pi}, \Blangle \bm{\pi}''-\bm{\pi}', \E_{\substack{s'\sim P}}[\delta_f(s,a,s')]\Brangle_a\Brangle_s + \Blangle \bm{d}^{\pi'} - \bm{d}_m^{\pi}, \E_{\substack{a\sim \pi \\ s'\sim P}}[\delta_f(s,a,s')]\Brangle_s \\ 
        \numberthis\label{e_td_pi''}
        & \qquad \qquad + \Blangle \bm{d}_m^{\pi}, \Blangle \bm{\pi}'' - \bm{\pi}', \E_{s'\sim P}[\delta_f(s,a,s')]\Brangle_a \Brangle_s + \Blangle \bm{d}_m^{\pi}, \E_{\substack{a\sim \pi \\ s'\sim P}}[\delta_f(s,a,s')] \Brangle_s \\
    \numberthis \label{e_td_pi'}
    \E_{\substack{s\sim d^{\pi} \\ a\sim \pi \\ s' \sim P}}[\delta_{f}(s,a,s')]
    &= \Blangle \bm{d}^{\pi} - \bm{d}_m^{\pi}, \E_{\substack{a\sim \pi \\ s'\sim P}}[\delta_f(s,a,s')]\Brangle_s + \Blangle \bm{d}_m^{\pi}, \E_{\substack{a\sim \pi \\ s'\sim P}}[\delta_f(s,a,s')] \Brangle_s 
\end{align*}
Inserting (\ref{e_td_pi''}) and (\ref{e_td_pi'}) into (\ref{ret_diff}) yields
\begingroup
\begin{align*} 
    J(\pi')-J(\pi)
    &= \frac{1}{1-\gamma}\big(\E_{\substack{s\sim d^{\pi'} \\ a\sim \pi' \\ s' \sim P}}[\delta_{f}(s,a,s')]-\E_{\substack{s\sim d^{\pi} \\ a\sim \pi \\ s' \sim P}}[\delta_{f}(s,a,s')]\big)\\
    \numberthis \label{ret_diff_ip1}
    &= 
            \frac{1}{1-\gamma} \big (
            \Blangle \bm{d}_m^{\pi}, \Blangle \bm{\pi}'' - \bm{\pi}', \E_{s'\sim P}[\delta_f(s,a,s')]\Brangle_a \Brangle_s \\
            & \qquad \qquad + \underbrace{
                \Blangle \bm{d}^{\pi'} - \bm{d}_m^{\pi}, \Blangle \bm{\pi}''-\bm{\pi}', \E_{\substack{s'\sim P}}[\delta_f(s,a,s')]\Brangle_a\Brangle_s
                }
                _{_\text{\mbox{\normalsize $\mathcal{E}_{\Delta\pi \Delta m}$}}} \\
            & \qquad \qquad + \underbrace{\Blangle \bm{d}^{\pi'} - \bm{d}^{\pi}, \E_{\substack{a\sim \pi\\s'\sim P}}[\delta_f(s,a,s')]\Brangle_s \big )}_{_\text{\mbox{\normalsize $\mathcal{E}_{\Delta\pi}$}}}
\end{align*}
\endgroup

The error terms in \ref{ret_diff_ip1} are named according to the distribution mismatch from which they arise: $\mathcal{E}_{\Delta\pi \Delta m}$ originates from a mismatch between state distributions collected under different policies and state transition functions, whereas $\mathcal{E}_{\Delta\pi}$ consists only of a policy-induced distribution mismatch. In order to relate these error terms to useful measures of policy- and transition probability divergences, Hölder's inequality for $L^p$ spaces is used extensively.

\begin{remark}
Hölder's inequality states that for any measurable space $\mathbb{S}$, $p,q\in[1,\infty]$ with the property $\frac{1}{p}+\frac{1}{q}=1$, and all real- or complex-valued functions $f$ and $g$ on $\mathbb{S}$ the following inequality holds:
\begin{align}\label{holder}
  \|fg\|_1\leq\|f\|_p\|g\|_q.
\end{align}
\end{remark}
Applying Hölder's inequality to the inner product on the vector space in $\mathbb{R}^{|\mathcal{A}|}$ straightforwardly gives 
\begin{align*} \label{pi_diff}
    \Blangle \bm{\pi}''-\bm{\pi}', \E_{\substack{s'\sim P}}[\delta_f(s,a,s')]\Brangle_a 
    & = \sum_{a} (\pi'(a|s)-\pi(a|s)) \E_{\substack{s'\sim P}}[\delta_f(s,a,s')] \\ 
    &\lesseqgtr \pm \sum_{a} |(\pi'(a|s)-\pi(a|s)) \E_{\substack{s'\sim P}}[\delta_f(s,a,s')]| \\ 
    &= \pm \| (\pi'(a|s)-\pi(a|s)) \E_{\substack{s'\sim P}}[\delta_f(s,a,s')] \|_1 \\
    & \lesseqgtr \pm \| \pi'(a|s)-\pi(a|s)\|_p \|\E_{\substack{s'\sim P}}[\delta_f(s,a,s')] \|_q \\
    & \lesseqgtr \pm 2D_{TV}(\pi'\|\pi) \max_a | \E_{\substack{s'\sim P}}[\delta_f(s,a,s')]| \,. \numberthis
\end{align*}
With $p=1$ follows $\|\pi'(a|s)-\pi(a|s)\|_1=2D_{TV}(\pi'\|\pi)$ and similarly setting $q=\infty$ yields $\|\E_{s'\sim P}[\delta_f(s,a,s')]\|_\infty=\max_{a}|\E_{s' \sim P} [ \delta_f(s,a,s')]|$. Before applying a similar approach to the error terms $\mathcal{E}_{\Delta\pi \Delta m}$ and $\mathcal{E}_{\Delta\pi}$, it is crucial to find an expression for the $L_1$ norm of the vector distance in state distributions under shifted policies and state transition dynamics. Recall, that the state distribution under policy $\pi'(a|s)$ and model dynamics can be written in vector form $\bm{d}^{\pi'}_m$ according to (\ref{state_dist}):
\begin{align*}
    \bm{d}^{\pi}_m = (1-\gamma)(\bm{I}-\gamma \bm{P}^{\pi}_m)^{-1} \bm{\mu}.
\end{align*}
Following Achiam et al.\ \cite{achiamConstrainedPolicyOptimization2017a}, the matrices $\bm{G}$ and $\bm{\Delta}$ are introduced to allow the reshaping
\begin{equation}
\begin{aligned}
    & \bm{G}^{\pi} = (\bm{I}-\gamma \bm{P}^{\pi})^{-1}     \qquad
    & & \bm{\Delta}^{\pi} = \bm{P}^{\pi'}-\bm{P}^{\pi}      \qquad
    & & \bm{d}^{\pi} =(1-\gamma) \bm{G}^{\pi}\bm{\mu}\\     \qquad
    & \bm{G}^{\pi'} = (\bm{I}-\gamma \bm{P}^{\pi'})^{-1}    \qquad
    & & \bm{\Delta}^{\pi}_m = \bm{P}^{\pi'}-\bm{P}^{\pi}_m  \qquad
    & & \bm{d}^{\pi'} =(1-\gamma) \bm{G}^{\pi'}\bm{\mu}\\   \qquad
    & \bm{G}^{\pi}_m = (\bm{I}-\gamma \bm{P}^{\pi}_m)^{-1}  \qquad
    & & \qquad  
    & &\bm{d}^{\pi}_m =(1-\gamma) \bm{G}^{\pi}_m\bm{\mu}    \qquad
\end{aligned}
\end{equation}
\begin{equation}
\begin{aligned}
    \bm{G}^{\pi-1}_m-\bm{G}^{\pi'-1}
    &= \gamma\bm{\Delta}^\pi_m \\
    \bm{G}^{\pi'}\bm{G}^{\pi-1}_m\bm{G}^{\pi}_m-\bm{G}^{\pi'}\bm{G}^{\pi'-1}\bm{G}^{\pi}_m
    &= \gamma \bm{G}^{\pi'} \bm{\Delta}_m^\pi \bm{G}^{\pi}_m \\ 
    \bm{G}^{\pi'}-\bm{G}^{\pi}_m
    &= \gamma \bm{G}^{\pi'} \bm{\Delta}_m^\pi \bm{G}^{\pi}_m \, \, .
\end{aligned}
\end{equation}
Accordingly, the distance of vectorized state distributions is described by
\begin{equation}\label{diff_s_dist_mat}
\begin{aligned}
    \bm{d}^{\pi'}-\bm{d}^{\pi}_m
    &= (1-\gamma)(\bm{G}^{\pi'}-\bm{G}^{\pi}_m)\bm{\mu} \\
    &= \gamma(1-\gamma)\bm{G}^{\pi'}\bm{\Delta}_m^\pi \bm{G}^{\pi}_m\bm{\mu} \\
    &= \gamma \bm{G}^{\pi'}\bm{\Delta}_m^\pi \bm{d}^{\pi}_m \, .    
\end{aligned}
\end{equation}
The $L_1$ norm of this vector difference is bounded by
\begin{equation} \label{stdist_diff_l1}
\begin{aligned}
    \|\bm{d}^{\pi'}-\bm{d}^{\pi}_m\|_1
    &= \| \gamma \bm{G}^{\pi'}\bm{\Delta}_m^\pi \bm{d}^{\pi}_m \|_1 \\
    &\leq \gamma \| \bm{G}^{\pi'} \|_1 \| \bm{\Delta}_m^\pi \bm{d}^{\pi}_m \|_1 \\
    &= \gamma \| (\bm{I}-\gamma \bm{P}^{\pi'})^{-1} \|_1 \| \bm{\Delta}_m^\pi \bm{d}^{\pi}_m \|_1 \\
    &\leq \gamma \big( \sum_{t=0}^\infty \gamma^t \| \bm{P}^{\pi'} \|_1^t \big) \| \bm{\Delta}_m^\pi \bm{d}^{\pi}_m \|_1 \\
    &= \frac{1}{1-\gamma} \| \bm{\Delta}_m^\pi \bm{d}^{\pi}_m \|_1 \,.
\end{aligned}
\end{equation}
Recall that $\bm{\Delta}_m^\pi = \bm{P}^{\pi'}-\bm{P}^{\pi}_m$ with entries $P^{\pi}_{ij} = \sum_a P(s'_j|s_i,a)\pi(a|s_i)$ and $P^{\pi}_{m, ij} = \sum_a P_m(s'_j|s_i,a)\pi(a|s_i)$ such that $ \bm{\Delta}_m^\pi \bm{d}^{\pi}_m$ is a vector-valued quantity in $\mathbb{R}^{|\mathcal{S}|}$. For finite action spaces $\mathcal{A}$, its $L_1$ norm thus becomes
\begin{equation}\label{stdist_diff}
\begin{aligned}
    \| \bm{\Delta}_m^\pi \bm{d}^{\pi}_m \|_1 
    &= \sum_{s'}\babs \sum_s\Delta_\pi^m(s'|s)d^{\pi}_m(s)\babs \\
    &\leq \sum_{s'}\sum_s\babs\Delta_\pi^m(s'|s)d^{\pi}_m(s)\babs \\
    &=\sum_{s}d^{\pi}_m(s)\sum_{s'}\babs\Delta_\pi^m(s'|s)\babs \\
    &= \sum_{s}d^{\pi}_m(s)\sum_{s'}\babs \sum_a P(s'|s,a)\pi'(a|s)-P_m(s'|s,a)\pi(a|s)\babs \\
    &= \sum_{s}d^{\pi}_m(s)\sum_{s'}\babs \sum_a P(s'|s,a)\big(\pi'(a|s)-\pi(a|s)\big)+ \pi(a|s)\big(P(s'|s,a)-P_m(s'|s,a)\big)\babs \\
    &\leq \sum_{s}d^{\pi}_m(s)\sum_{s'} \sum_a P(s'|s,a) \babs \pi'(a|s)-\pi(a|s)\babs + \, \pi(a|s)\babs P(s'|s,a)-P_m(s'|s,a)\babs \\
    &= \sum_{s}d^{\pi}_m(s) \sum_a  \babs \pi'(a|s)-\pi(a|s)\babs  \sum_{s'}P(s'|s,a) +\sum_s d^{\pi}_m(s) \sum_a \pi(a|s)\sum_{s'}\babs P(s'|s,a)-P_m(s'|s,a)\babs \\
    &= \E_{s\sim d^{\pi}_m}\big[2D_{TV}(\pi'(a|s)\|\pi(a|s))\big] + \E_{\substack{s\sim d^{\pi}_m \\ a\sim \pi}} \big[ 2D_{TV}(P(s'|s,a)\|P_m(s'|s,a))\big] \,.
\end{aligned}    
\end{equation}
Inserting (\ref{stdist_diff}) in (\ref{stdist_diff_l1}) yields the important interim result given by Lemma\ \ref{stdist_diff_lemma}.
\begin{lemma} \label{stdist_diff_lemma}
    Let $\pi(a|s)$ and $\pi'(a|s)$ be arbitrary policies on finite state and action spaces $s \in \mathcal{S}$, $a \in \mathcal{A}$. Then for state transition probabilities $P(s'|s,a)$ and $P_m(s'|s,a)$, the $L_1$ norm of the vector-valued difference in stationary state distributions $\| \bm{d}^{\pi'}-\bm{d}^{\pi}_m \|_1$ is upper bounded by the identity
    \begin{align*}+
        \|\bm{d}^{\pi'}-\bm{d}^{\pi}_m\|_1
        &\leq \frac{2 \gamma}{1-\gamma} \big( \E_{s\sim d^{\pi}_m}\big[D_{TV}(\pi'(a|s)\|\pi(a|s))\big] + \E_{\substack{s\sim d^{\pi}_m \\ a\sim \pi}} \big[ D_{TV}(P(s'|s,a)\|P_m(s'|s,a))\big] \big) \,,
    \end{align*}
    where, $D_{TV}(p\|q)$ is the total variation distance between $p$ and $q$.
\end{lemma}
% For the case $P(s'|s,a) = P_m(s'|s,a)$, Lemma \ref{stdist_diff_lemma} gives
% \begin{align} \label{stdist_diff_pionly}
%     \|\bm{d}^{\pi'}-\bm{d}^{\pi}\|_1
%         &\leq \frac{2 \gamma}{1-\gamma} \big(\E_{s\sim d^{\pi}}\big[D_{TV}(\pi'(a|s)\|\pi(a|s))\big] \big) \, .
% \end{align}
Regrouping the results (\ref{stdist_diff}) and (\ref{pi_diff}) is now helpful in further breaking down the error terms $\mathcal{E}_{\Delta\pi \Delta m}$ and $\mathcal{E}_{\Delta\pi}$ described in (\ref{ret_diff_ip1}). Again by employing Hölder's inequality, we obtain
\begingroup
%\allowdisplaybreaks
\begin{align*}
    \mathcal{E}_{\Delta \pi \Delta m} 
    & = \Blangle \bm{d}^{\pi'} - \bm{d}_m^{\pi}, \Blangle \bm{\pi}''-\bm{\pi}', \E_{\substack{s'\sim P}}[\delta_f(s,a,s')]\Brangle_a \Brangle_s \\
    & \lesseqgtr \Blangle \bm{d}^{\pi'} - \bm{d}_m^{\pi},
        \quad \pm 2D_{TV}(\pi'\|\pi) \max_a | \E_{\substack{s'\sim P}}[\delta_f(s,a,s')]| \Brangle_s \\
    \numberthis \label{eps_pim_A}
    & \lesseqgtr \pm \| \bm{d}^{\pi'} - \bm{d}_m^{\pi}\|_1 
        \| 2D_{TV}(\pi'\|\pi) \max_a | \E_{\substack{s'\sim P}}[\delta_f(s,a,s')]| \, \|_\infty \\
    & \lesseqgtr \pm \frac{4\gamma}{1-\gamma} 
    \max_s \babs D_{TV}(\pi'\|\pi) \max_a \E_{\substack{s'\sim P}}[\delta_f(s,a,s')]\babs 
    \big(
        \E_{s\sim d^{\pi}_m}\big[D_{TV}(\pi'\|\pi)\big] 
        + \E_{\substack{s\sim d^{\pi}_m \\ a\sim \pi}} \big[ D_{TV}(P\|P_m)\big] 
        \big) \,,
    \\
    \mathcal{E}_{\Delta \pi}
    & = \Blangle \bm{d}^{\pi'} - \bm{d}^{\pi}, \E_{\substack{a\sim \pi \\ s'\sim P}}[\delta_f(s,a,s')]\Brangle_s \\
    \numberthis \label{eps_pi_A}
    & \lesseqgtr \pm \| \bm{d}^{\pi'} - \bm{d}^{\pi}\|_1 
        \| \E_{\substack{a\sim \pi \\ s'\sim P}}[\delta_f(s,a,s')] \|_\infty \\
    & \lesseqgtr \pm \frac{2\gamma}{1-\gamma} \big( 
        \E_{\substack{s\sim d^{\pi}}} \big[ D_{TV}(\pi'\|\pi)\big] 
        \big)
        \max_s \babs \E_{\substack{a \sim \pi \\ s'\sim P}}[\delta_f(s,a,s')] \babs \,.
\end{align*}
% Due to concatenated use of Hölder's inequality, the term $\delta_f(s,a,s')$ is present under the maximum over actions in $\mathcal{E}_{\Delta \pi \Delta m}$ and as an expected value in $\mathcal{E}_{\Delta \pi}$. While the maximum term $\max_a | \E_{\substack{s'\sim P}}[\delta_f(s,a,s')]|$ still requires evaluation under the true next-state expectation, we found further upper bounding to be fruitless since we were unable to eliminate the non-zero maximum term on $\max_a | \E_{\substack{s'\sim P_m}}[\delta_f(s,a,s')]|$. 
We can derive a tighter expression for the term $\max_s | \E_{a \sim \pi, s'\sim P}[\delta_f(s,a,s')] |$ by relating the expression to next-state distributions according to $s'\sim P_m$ and receive an additional factor in $D_{TV}(P \| P_m)$. For this, we rewrite
% For this, let $\epsilon_{\Delta m \delta}(s,a)$ be the one-step error associated with the difference in vector-valued state transition probabilities $\bm{P}(s,a) \in \mathbb{R}^{|\mathcal{S}|}$, $\bm{P}_m(s,a) \in \mathbb{R}^{|\mathcal{S}|}$, and the vector form of $\delta_f(s,a,s')$ given by $\bm{\delta}_f(s,a)\in \mathbb{R}^{|\mathcal{S}|}$. 
% \begin{align}
%     \epsilon_{\Delta m \delta} = \Blangle \bm{P}(s,a)-\bm{P}_m(s,a), \bm{\delta}_f(s,a)\Brangle_{s'}.
% \end{align} 
% Then, the expected value of the temporal difference residual over actions and successor states can be written as
\begin{equation} \label{exp_s'_delta_f_A}
    \begin{split}
        \E_{\substack{a \sim \pi \\ s'\sim P}}[\delta_f(s,a,s')] 
        & = \E_{a \sim \pi} \big [ 
            \Blangle \bm{P}(s,a)-\bm{P}_m(s,a),\bm{\delta}_f(s,a)\Brangle_{s'} + \E_{s'\sim P_m} \big[ \delta_f(s,a,s') \big] 
        \big ]\\
        & = \E_{a \sim \pi} \big [ \epsilon_{\Delta m \delta}(s,a) \big ]
            + \E_{\substack{a \sim \pi \\ s'\sim P_m}} \big[ \delta_f(s,a,s') \big] \\
        & \leq
            \E_{a \sim \pi} \big[ 2 D_{TV}(P\|P_m) \max{s'}\babs \delta_f(s,a,s') \babs \big]
            + \E_{\substack{a \sim \pi \\ s'\sim P_m}} \big[ \delta_f(s,a,s') \big]\,.
    \end{split}
\end{equation}
The value of this reshaping becomes apparent by making a choice for the still arbitrary function $f(s)$. Note that for $f(s) = V^{\pi}_m(s)$, the expression $\E_{s' \sim P_m} [\delta_{V_m^{\pi}}(s,a,s')] = \E_{s' \sim P_m} [r(s,a,s') + \gamma V_m^{\pi}(s') - V_m^{\pi}(s)]$ becomes the advantage function of policy $\pi$ under model dynamics. Then, per definition of the advantage function, $\E_{a\sim \pi}[A_{\pi}^m(s,a)]=0$ can be leveraged. Accordingly, we set $f(s) = V^{\pi}_m(s)$ to obtain
% However, it should be pointed out that this assumption may not hold true for approximations $\tilde{V}_m^{\pi}(s)$, which would result in an additional error factor on $D_{TV}(\pi' \| \pi)$. With $f(s) = V_m^{\pi}(s)$ and \ref{exp_s'_delta_f_A}, the results from \ref{eps_pim_A} and \ref{eps_pi_A} can further be bounded.
\begingroup
%\allowdisplaybreaks
\begin{align*}
    \mathcal{E}_{\Delta \pi \Delta m} 
    & \lesseqgtr \pm \frac{4\gamma}{1-\gamma} \max_{s,a,s'} \babs \delta_{V_m^{\pi}}(s,a,s') \babs 
    \Big(
        \max_{s}D_{TV}(\pi'||\pi)^2
        + \max_{s}D_{TV}(\pi'||\pi)\max_{s,a}D_{TV}(P||P_m)
        \Big) \\
    \numberthis \label{eps_delta_pim2_A}
    &= \pm \frac{4 \gamma}{1-\gamma} \delta_{V_m^{\pi}}^{\max} \big(
        \epsilon_{\pi}^2
        + \epsilon_{\pi} \epsilon_{m}
        \big)\\ 
    \\
    \mathcal{E}_{\Delta \pi}
    & \lesseqgtr \pm \frac{2\gamma}{1-\gamma} \big( 
        \E_{\substack{s\sim d^{\pi}}} \big[ D_{TV}(\pi'||\pi)\big] 
        \big) 
        \max_s \Babs 
            \E_{a \sim \pi} \big[ 2 D_{TV}(P||P_m) \max_{s'}\babs \delta_{V_m^{\pi}}(s,a,s') \babs \big]
        \Babs \\
    & \lesseqgtr \pm 
        \frac{4\gamma}{1-\gamma} \big( 
        \max_s D_{TV}(\pi'||\pi) \max_{s,a} D_{TV}(P||P_m)
        \max_{s,a,s'}\babs \delta_{V_m^{\pi}}(s,a,s')\babs
        \big) \\
    \numberthis \label{eps_delta_pi2_A}
    &= \pm 
        \frac{4\gamma}{1-\gamma} \delta_{V_m^{\pi}}^{\max} \big( 
        \epsilon_{\pi} \epsilon_{m}
        \big) 
        \\ \\ 
    & \text{where} \\ \\
    \epsilon_{\pi} 
    &= D_{TV}^{max}(\pi' \| \pi) =\max_s D_{TV}(\pi'(a|s) \| \pi(a|s)) \\
    \epsilon_m 
    &= D_{TV}^{max}(P \| P_m)=\max_{s,a} D_{TV}(P(s'|s,a) \| P_m(s'|s,a)) \\
    \delta_{V_m^{\pi}}^{\max}
    &= \max_{s,a,s'} \babs \delta_{V_m^{\pi}}(s,a,s') \babs \,.
\end{align*}
Regrouping these results with the original expression for return differences in (\ref{ret_diff_ip1}), we get
\begin{equation}\label{ret_diff_ip3_A}
\begin{aligned}
\Delta J_{\pi}(\pi')
    &= J(\pi')-J(\pi) \\
    &\lesseqgtr 
        \frac{1}{1-\gamma} \big (
            \Blangle \bm{d}_m^{\pi}, \Blangle \bm{\pi}'' - \bm{\pi}', \E_{s'\sim P}[\delta_{V_m^{\pi}}(s,a,s')]\Brangle_a \Brangle_s
            + \mathcal{E}_{\Delta \pi \Delta m} 
            + \mathcal{E}_{\Delta \pi} 
            \big ) \\
    &=  
        \frac{1}{1-\gamma} \Big (
                \Blangle \bm{d}_m^{\pi}, \Blangle \bm{\pi}'' - \bm{\pi}', \E_{s'\sim P}[\delta_{V_m^{\pi}}(s,a,s')]\Brangle_a \Brangle_s
                \pm \frac{4 \gamma}{1-\gamma} \delta_{V_m^{\pi}}^{\max} \big(
                    \epsilon_{\pi}^2
                    + 2 \epsilon_{\pi} \epsilon_{m}
                    \big)
                    \Big ) \, . 
\end{aligned}    
\end{equation}
Finally, the expected value for the one step transitions in $\E_{s'\sim P}[\delta_{V_m^{\pi}}(s,a,s')]$ is related to a model transition distribution using (\ref{exp_s'_delta_f_A}), and the inner product over $\langle \bm{\pi'}-\bm{\pi}, \bm{\cdot} \rangle_a$ is replaced by its importance sampling ratio. The resulting expression $\E_{s'\sim P_m}[\delta_{V_m^{\pi}}(s,a,s')]=A_m^{\pi}(s,a)$ recovers the definition of the model-advantage function with
\begin{equation}\label{onesteptderr}
\begin{aligned}
    \Blangle \bm{\pi}'' - \bm{\pi}', \E_{s'\sim P}[\delta_{V_m^{\pi}}(s,a,s')]\Brangle_a 
    &\lesseqgtr  \E_{\substack{a \sim \pi \\s' \sim P_m}} \big[\big( \frac{\pi'}{\pi}-1\big) \delta_{V_m^{\pi}}(s,a,s') \big] \pm 4\epsilon_m \epsilon_\pi \delta_{V_m^{\pi}}^{\max} \\
    &= \E_{a \sim \pi} \big[ \big( \frac{\pi'}{\pi} \big) A_m^{\pi}(s,a) \big] \pm 4\epsilon_m \epsilon_\pi \delta_{V_m^{\pi}}^{\max}.
\end{aligned}
\end{equation}
Boundary \ref{boundary1} is obtained by combining\ (\ref{ret_diff_ip3_A}) and\ (\ref{onesteptderr});
% Regrouping the identities\ \ref{ret_diff_ip3_A},\ \ref{onesteptderr}, and assigning more practical names yields Boundary \ref{boundary1}.
\begin{equation}
\begin{aligned} 
    & \Delta J(\pi, \pi')
    \lesseqgtr \frac{L^{\pi}_m(\pi')}{1-\gamma}
        \pm \frac{4 \delta^{max} \epsilon}{1-\gamma} \\
& \text{where} \\
    & L^{\pi}_m(\pi')
    =  \E_{\substack{s \sim d_{m}^{\pi} \\ a \sim \pi}}
        \big[
            \big( \frac{\pi'(a|s)}{\pi(a|s)}\big) A^{\pi}_m(s,a)
        \big]\\
    & \delta^{max} = \max_{s,a,s'} \big| r(s,a,s') + V^{\pi}_{m}(s') - V^{\pi}_{m}(s) \big| \\
    & \epsilon = \epsilon_{\pi}\epsilon_m
        + \frac{\gamma}{1-\gamma}
        (
            \epsilon_{\pi}^2
            + 2 \epsilon_{\pi} \epsilon_{m}
        )\\
    & \epsilon_{\pi}
    = \max_{s}D_{TV}(\pi'(\cdot|s)||\pi(\cdot|s)) \\
    & \epsilon_{m}
    = \max_{s,a}D_{TV} \big(P(\cdot|s,a)||P_m(\cdot|s,a)\big) \,.
\end{aligned}    
\end{equation}

% \subsubsection{Policy Advantage, Kakade and Langford} \label{kakadeproof}
% Proof for the difference in discounted returns between policies $J(\pi')-J(\pi)$ is provided. We follow the proof provided by Kakade and Langford\ \cite{kakadeApproximatelyOptimalApproximate2002a} and the summary given by Schulman et al.\ \cite{schulmanTrustRegionPolicy2015}. In the following, the notation $s_t = s$, $a_t=a$, $s_{t+1}=s'$ is used in the infinite horizon case for simplicity. By leveraging the definition of the advantage function $A^\pi(s,a)=\E_{s' \sim P} [r(s,a,s') + \gamma V^\pi(s')-V^\pi(s)]$ and the discounted returns $J(\pi)=\E_{s_0 \sim \mu}[V^\pi(s_0)]$, the return differences $J(\pi')-J(\pi)$ can be rewritten as 
% \begin{align*}
%     J(\pi')-J(\pi) 
%     &= \E_{\tau \sim \pi'}\big[ \sum_{t=0}^\infty \gamma^t r(s,a,s')\big] - \E_{s_0 \sim \mu}[V^\pi(s_0)] \\
%     &= \E_{\tau \sim \pi'}\big[-V^\pi(s_0) + \sum_{t=0}^\infty \gamma^t r(s,a,s')\big] \\
%     &= \E_{\tau \sim \pi'}\big[ \sum_{t=0}^\infty \gamma^t \big( r(s,a,s')+\gamma V^\pi(s_{t+1}) - V^\pi(s_t)) \big] \\ \numberthis
%     &= \E_{\tau \sim \pi'}\big[ \sum_{t=0}^\infty \gamma^t A^\pi(s_t,a_t) \big] \\
% \end{align*}
% where the continuity property of $V^\pi(s_0) = \sum_{t=0}^\infty \gamma^t V^\pi(s_t) - \gamma^{t+1} V^\pi(s_{t+1})$ was used.
\clearpage
\section*{APPENDIX B}
\subsection{Experimental Setup}
\subsubsection{Reward and Cost Functions}
\paragraph{AntSafe} The reward function is equal to the rewards in the common \textit{Ant-v2} environment and comprises the torso velocity in global x-direction, a negative control reward on exerted torque, a negative contact reward and a constant positive reward for survival, which results in
\begin{align*}
    r = &\frac{x_{\textit{torso},t+1} - x_{\textit{torso},t}}{dt}
      - \frac{1}{2} \big\| \bm{a_t} \big\|^2_2 
      - \frac{1}{2*10^3} \big\| \textbf{\textit{contact}}_t \big \|^2_2  
      + 1 \,,
\end{align*}
where $x_{\textit{torso},t}$ is the x-position in global coordinates, $\bm{a_t}$ is the vector of chosen actions, and $\textbf{\textit{contact}}_t$ are contact forces registered by the mujoco simulation. The cost function for this environment is defined by an agent's proximity to the maze walls and its survival. The agent receives a cost of $1$ for every timestep in which his torso's distance to a maze wall is below a threshold of $1.8$. In addition, a cost of $1$ is given for a terminal state, that is when the agent has caused the robot to topple over.
\begin{align*}
    c = 
    \begin{cases}
    0, & \text{for} \quad 0.2 \leq z_{\textit{torso}, t+1} \leq 1.0\\
    & \text{and} \quad \big\| \bm{x}_{\textit{torso}, t+1} - \bm{x}_{wall} \big\|_2 \geq 1.8 \\
    1, & \text{else} \,.
    \end{cases}
\end{align*}
Here, $z_{\textit{torso}, t+1}$ is the robot's torso-coordinate in z-direction, $\bm{x}_{\textit{torso}, t+1}$ is the x,y,z-vector of the torso, and $\bm{x}_{wall}$ is the vector to the closest point of a wall. 
\paragraph{HalfCheetahSafe} The reward function in HalfCheetahSafe is composed of the torso velocity in global x-direction and a negative control reward on exerted torque, equal to the common \textit{HalfCheetah-v2} environment. $r$ is thus defined as 
\begin{align*}
    r = &\frac{x_{\textit{torso},t+1} - x_{\textit{torso},t}}{dt}
      - \frac{1}{2} \big\| \bm{a}_t \big\|^2_2 \,.
\end{align*}
In addition, agents incur costs for unsafe proximity to the heading object according to 
\begin{align*}
    c = 
    \begin{cases}
    0, & \text{for} \quad |x_{\textit{torso},t+1} - x_{\textit{obj,t}}| \geq 2.0 \\
    1, & \text{else} . 
    \end{cases}
\end{align*}
Again, $x_{\textit{torso},t}$ is the torso's x-position in global coordinates, and $\bm{a}_t$ the control inputs by the agent.

\clearpage

\subsubsection{Hyperparameter settings for CMBPO}
\hfill
\begin{table}[!ht]
    \centering
    \renewcommand{\arraystretch}{1.1}
    \small    
    \centering
        \begin{tabularx}{\textwidth}{l|X|X|X}
         Hyperparameter & AntSafe & HalfCheetahSafe & AntCircle\\
         \hline
         \multicolumn{4}{c}{Policy and Value Learning} \\
         \hline
         Cost limit & 10 & 10 & 10 \\
         Policy network hidden layers & (256, 256) & (256, 256) &  (256, 256)\\
         Value network hidden layers & (128, 128) & (128, 128) & (128, 128)\\
         discount $\gamma$ (ret;cost) & 0.99; 0.97 & 0.99; 0.97 & 0.99; 0.97 \\
         GAE discount $\lambda$ (ret;cost)& 0.95; 0.5 & 0.95; 0.5 & 0.95; 0.5\\
         Target KL & 0.01 & 0.01 & 0.01\\
         Policy batch size & 50000 & 50000 & 50000 \\
         Conjugate gradient steps & 10 & 10 & 10 \\
         Conjugate gradient damping & 0.1 & 0.1 & 0.1 \\
         Backtracking steps & 10 & 10 & 10\\
         Entropy regularization & 0 & 0 & 0 \\
         Value learn rate & $3\times10^{-4}$ &$3\times10^{-4}$ & $3\times10^{-4}$\\
         Value mini-batch size & 2048 & 2048 & 2048\\
         Value train repeats & 8 & 8 & 8 \\
         \hline
         \multicolumn{4}{c}{Model Learning} \\
         \hline
        Dynamics model hidden layers & (512,512) & (512,512) & (512,512) \\
        Ensemble size & 7 & 7 & 7 \\
        Ensemble elites & 5 & 5 & 5\\
        Learn rate & 0.001 & 0.001 & 0.001 \\
        Training batch size & 2048 & 2048 & 2048\\
        Initial ratio of real samples $\alpha_0$ & 0.4 & 0.3 & 0.2 \\
        Initial rollout length $h_0$ & 5 & 5 & 7 \\
        Start-state sampling temperature $\beta$ & 2 & 2 & 2 \\
        \hline
        \end{tabularx}
    \label{tab:hpas}
    \caption{Hyperparameter Settings for AntSafe, HalfCheetahSafe, and AntCircle}
\end{table}
\clearpage
\end{document}